\documentclass[letterpaper]{article}
\usepackage{aaai25}
\usepackage{times}
\usepackage{helvet}
\usepackage{courier}
\usepackage[hyphens]{url}
\usepackage{graphicx}
\usepackage{natbib}
\usepackage{caption}
\usepackage{algorithm}
\usepackage{algorithmic}

\usepackage{newfloat}
\usepackage{listings}
\DeclareCaptionStyle{ruled}{labelfont=normalfont,labelsep=colon,strut=off} %
\floatstyle{ruled}
\newfloat{listing}{tb}{lst}{}
\floatname{listing}{Listing}
\usepackage{booktabs}
\usepackage{amsmath}
\usepackage{amsfonts}
\usepackage{adjustbox}
\usepackage{longtable}
\usepackage{tabularx}
\usepackage{multirow}
\usepackage{subcaption}

\title{TSKANMixer: Kolmogorov–Arnold Networks with MLP-Mixer Model for Time Series Forecasting}
\author{
    Young-Chae Hong\textsuperscript{\rm 1},
    Bei Xiao\textsuperscript{\rm 1},
    Yangho Chen\textsuperscript{\rm 1}
}
\affiliations{
    \textsuperscript{\rm 1}Amazon\\
    Seattle, WA 98109\\
    hongych@amazon.com, bxxiao@amazon.com, yanghoc@amazon.com
}

\begin{document}

\maketitle

\begin{abstract}
Time series forecasting has long been a focus of research across diverse fields, including economics, energy, healthcare, and traffic management. Recent works have introduced innovative architectures for time series models, such as the Time-Series Mixer (TSMixer), which leverages multi-layer perceptrons (MLPs) to enhance prediction accuracy by effectively capturing both spatial and temporal dependencies within the data. In this paper, we investigate the capabilities of the Kolmogorov-Arnold Networks (KANs) for time-series forecasting by modifying TSMixer with a KAN layer (TSKANMixer). Experimental results demonstrate that TSKANMixer tends to improve prediction accuracy over the original TSMixer across multiple datasets, ranking among the top-performing models compared to other time series approaches. Our results show that the KANs are promising alternatives to improve the performance of time series forecasting by replacing or extending traditional MLPs.
\end{abstract}

\section{Introduction}
Time-series analysis is essential across a wide range of domains, including retail \cite{bose2017probabilistic}, finance \cite{taylor2008modelling}, economics \cite{granger2014forecasting}, transportation \cite{chen2001freeway, yin2021deep}, energy \cite{martin2010prediction, qian2019review, heidrich2020forecasting}, healthcare \cite{bui2018time, kaushik2020ai}, and climate \cite{wu2023interpretable}, where understanding and forecasting temporal patterns is crucial for decision-making and planning. In recent years, various deep learning (DL)-based forecasting models, including convolutional neural networks (CNNs), recurrent neural networks (RNNs), multi-layer perceptrons (MLPs), and Transformers, have been extensively studied to capture the complexity in real-world time-series datasets that are often multivariate with complex, non-linear dependencies among them \cite{wang2024deep, liu2024deep}.

However, contrary to the common intuition that DL-based models should be more effective than univariate models, it is shown that Transformer-based models can indeed be significantly worse than simple univariate temporal linear models on many commonly used forecasting benchmarks since they suffer from overfitting \cite{nie2022time, zeng2023transformers}. Instead, recent work has demonstrated that simple univariate linear models can outperform such deep learning models on several commonly used academic benchmarks. Recently, Chen et al. \cite{chen2023tsmixer}, inspired by the well-known MLP Mixer architecture in computer vision \cite{tolstikhin2021mlp}, proposed a fully MLP-based architecture for time series forecasting, Time-Series Mixer (TSMixer), by alternatively stacking multiple MLPs to capture temporal information in the time-domain and cross-variate information in the feature-domain. The authors showed that state-of-the-art performance can be achieved without necessarily relying on Transformers by demonstrating TSMixer's superior performance on benchmarks like the M5 dataset.

On the other hand, more recently, Kolmogorov-Arnold Networks (KANs) \cite{liu2024kan} was proposed  as a promising alternative to MLPs. Unlike traditional MLPs that have fixed activation functions on nodes, KANs utilize learnable activation functions on edges and perform instead a simple summation on nodes. The authors introduce KANs as a powerful new neural network architecture that can improve performance and interpretability compared to MLPs. This obviously opens opportunities for further improving deep learning models which rely heavily on MLPs \cite{liu2024kan}.

Recent research has explored the application of KANs for time-series. Xu et al. \cite{xu2024kolmogorov} investigated the use of KANs for time series forecasting and demonstrated that two KAN models significantly outperformed traditional forecasting methods. Similarly, Vaca-Rubio et al. \cite{vaca2024kolmogorov} showed that KANs outperformed conventional Multi-Layer Perceptrons (MLPs) in a real-world satellite traffic forecasting task, providing more accurate results with considerably fewer learnable parameters. Finally, Genet et al. \cite{genet2024tkan} proposed the adaptation of KANs to temporal sequences by combining recurrent neural networks (RNNs) and KANs. These researches confirm that the idea developed in the original KAN paper works well on real-world use cases and is highly relevant for time series analysis. In this paper, inspired by the KANs, we propose a new neural network architecture, TSKANMixer, by investigating the application of KANs to TSMixer for time series forecasting.

This paper is structured as follows. Section 2 presents the related work, providing fundamental background on KANs and TSMixer. Section 3 introduces the overall architecture of TSKANMixer, which uses a KAN layer in TSMixer. Computational experiments are presented in Section 4. Finally, conclusions are provided in Section 5.

\section{Related Work}

\subsection{Time-Series Mixer (TSMixer)}
\begin{figure}[htbp]
\centering
\includegraphics[width=0.9\columnwidth]{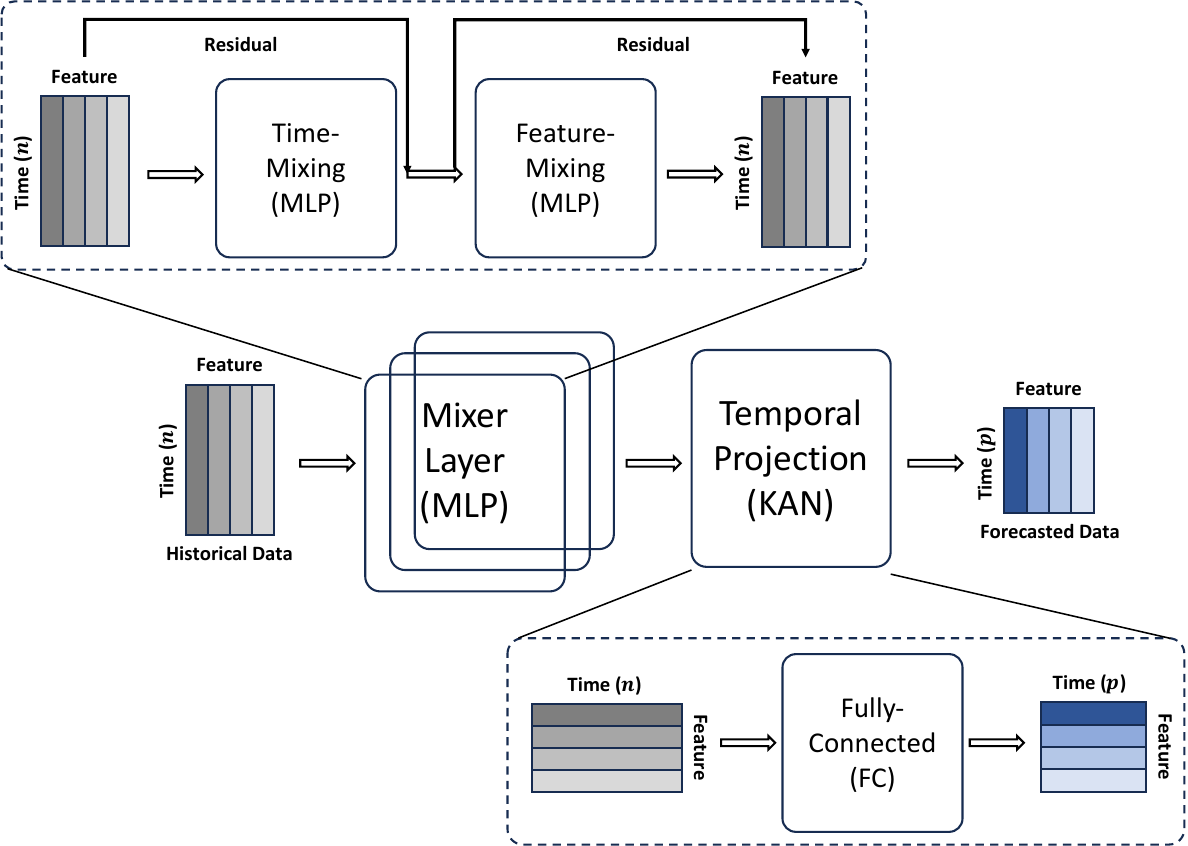}
\caption{TSMixer for multivariate time series forecasting \cite{chen2023tsmixer}}
\label{figure:TSMixer}
\end{figure}

TSMixer is an MLP-based architecture for time series forecasting \cite{chen2023tsmixer}, which analyzes the performance of linear models for time series forecasting rather than RNNs or Transformer-based frameworks and demonstrates its competitive performance on several time series forecasting benchmarks. TSMixer consists of multiple MLP layers across time and feature dimensions (i.e., time-mixing and feature-mixing MLP block) to capture time-domain temporal patterns and feature-domain cross-variate information alternatively with residual connections and batch norm. The residual designs ensure that TSMixer retains the capacity of temporal linear models. In contrast to recent Transformer-based models, the architecture of TSMixer is relatively simple to implement. Despite its simplicity, it demonstrates that TSMixer remains competitive with state-of-the-art models at representative benchmarks \cite{chen2023tsmixer}. The detail of TSMixer architecture is shown in Figure \ref{figure:TSMixer}.

\subsection{Kolmogorov-Arnold Network (KAN)}
As MLPs are based on the universal approximation theorem \cite{cybenko1989approximation}, which states that neural networks with a single hidden layer can approximate any continuous function with finite support, KANs rely on the Kolmogorov-Arnold representation theorem \cite{arnold2009functions, arnold2009representation}. The theorem states that any multivariate continuous function $f(x)$ on a bounded domain, where $x=(x_{1}, \dots, x_{n})$, can be written as a finite composition of continuous functions of a single variable and the binary operation of addition. Formally, a multivariate continuous function $f(x) : [0,1]^{n} \rightarrow \mathbb{R}$ can be represented by the finite composition of univariate functions \cite{liu2024kan}:

\begin{align}
f(x) = f(x_{1}, \dots , x_{n}) = \sum_{j=1}^{2n+1} \Phi_j \left( \sum_{i=1}^{n} \phi_{j,i} (x_i) \right) \label{eq:KAN}
\end{align}

where an outer function is $\Phi_j: \mathbb{R} \rightarrow \mathbb{R}$ and an inner function is $\phi_{j,i}: [0,1] \rightarrow \mathbb{R}$. 

As a MLP consists of layers where each layer performs a linear transformation followed by a non-linear activation function, a KAN layer can be defined as a matrix $\mathbf{\Phi}$ of univariate functions:

\begin{align}
\mathbf{\Phi(x)} = \{ \phi_{j,i} \}, \qquad i = \{1, \dots, n_{in}\}, \, j = \{1, \dots, n_{out}\} \label{eq:KAN_layer}
\end{align}

where the univariate functions $\phi_{j,i}$ have trainable parameters and $n_{in}$ is the number of inputs and $n_{out}$ is the number of outputs.

Generally, KANs can be expressed by a composition of multiple KAN layers, $y = \textbf{KAN}(x) = (\mathbf{\Phi_{L}} \circ \cdots \circ \mathbf{\Phi_{1}})(x)$ where $L$ is the number of layers. Then, the equation \ref{eq:KAN} for the Kolmogorov-Arnold representation theorem can be represented by a two-depth KAN layer of shape $[n, 2n + 1, 1]$, consisting of an inner layer with $n_{in} = n$ and $n_{out} = 2n + 1$, and an outer layer with $n_{in} = 2n + 1$ and $n_{out} = 1$ \cite{liu2024kan}.

While MLPs employ fixed activation functions on nodes, KANs employ learnable activation functions on edges \cite{liu2024kan}. Specifically, KANs learn activation patterns dynamically by replacing traditional linear weights on MLPs with univariate functions parameterized as splines, where a spline is defined by the order $k$ (the degree of the polynomial functions used to interpolate the curve between control points), and the number of intervals $G$ (the number of segments between adjacent control points). During spline interpolation, the control points separated by $G$ intervals are connected to form a smooth curve \cite{vaca2024kolmogorov}. Through learnable activation functions, KANs improve accuracy and interpretability while maintaining comparable or superior performance with more compact architectures across various tasks.

Vaca-Rubio et al. \cite{vaca2024kolmogorov} demonstrate that KANs consistently outperform MLPs with lower error metrics while achieving better results with reduced computational resources in time series forecasting. However, due to their intrinsic architecture, KANs have $(k + G)$ times more learnable parameters compared to MLPs \cite{yu2024kan}. To enhance computational efficiency, several regularization techniques have proven effective in optimizing KAN training \cite{cheon2024improving}. Specifically, the incorporation of dropout, weight decay, and batch normalization not only accelerates convergence but also significantly improves the model's generalization capabilities. Additionally, Bayesian optimization can be leveraged to reduce the parameter search space for more efficient training \cite{snoek2012practical}.

\section{TSKANMixer Architecture}
\begin{figure}[htbp]
\centering
\begin{subfigure}[b]{0.9\columnwidth}
   \centering
   \includegraphics[width=\textwidth]{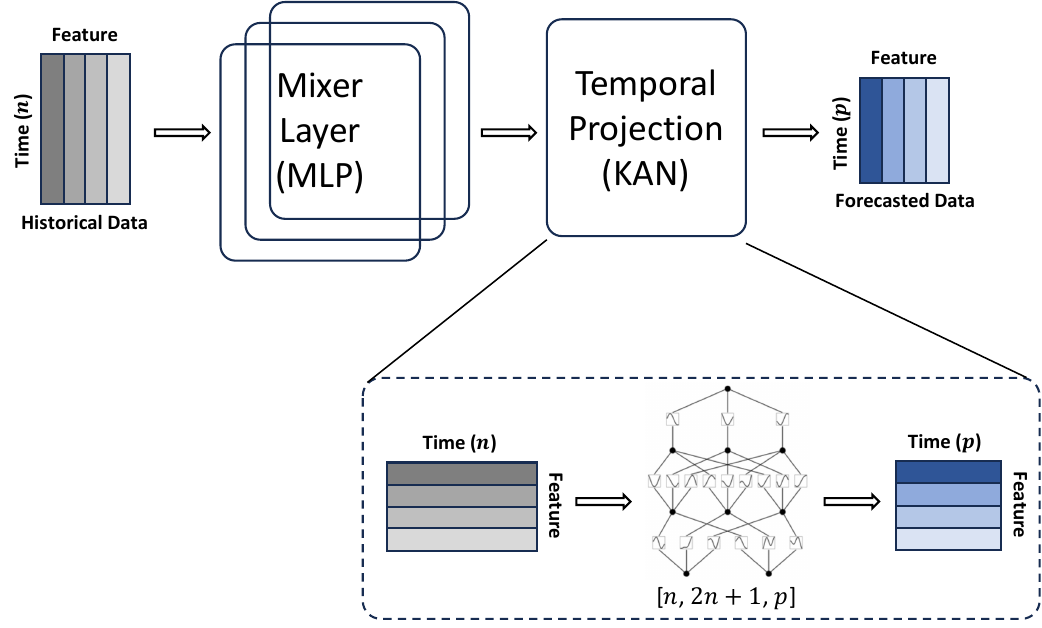}
   \caption{Version 01: Mixer Layer (=MLP) + Temporal Projection (=KAN)}
   \label{fig:TSKAN_01}
\end{subfigure}
\vskip\baselineskip
\begin{subfigure}[b]{0.9\columnwidth}
   \centering
   \includegraphics[width=\textwidth]{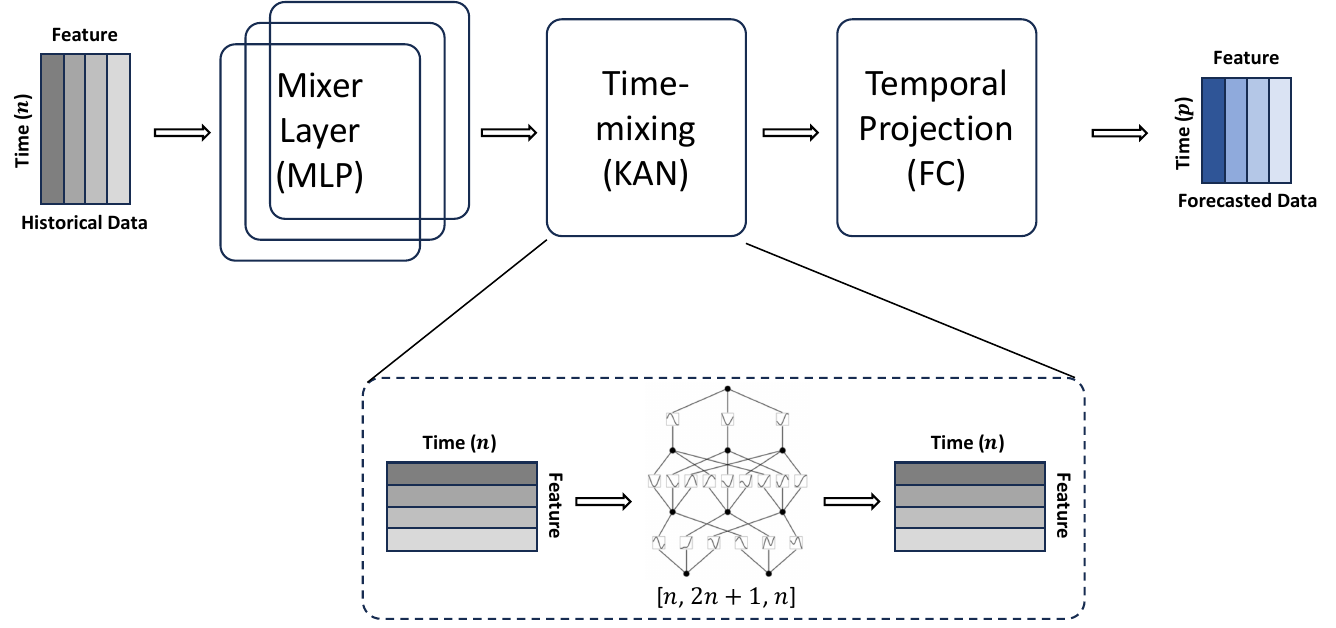}
   \caption{Version 02: Mixer Layer (=MLP) + Time-mixing Layer (=KAN) + Temporal Projection (=FC)}
   \label{fig:TSKAN_02}
\end{subfigure}
\caption{TSKANMixer Architectures}
\label{fig:TSKAN}
\end{figure}

In this paper, we explore and evaluate the application of a KAN layer to the MLP-based TSMixer architecture. We introduce two architectures of TSKANMixer as illustrated in Figure \ref{fig:TSKAN}. The proposed models apply the KAN framework to learn complex, non-linear relationships in temporal data. The first proposed architecture, presented in Figure \ref{fig:TSKAN_01}, uses KAN for temporal projection on the time domain as an alternative to a fully-connected layer in TSMixer \cite{chen2023tsmixer}. It maps the time series from the input length $L$ to the forecast horizon $H$ by learning the complex relationships between past inputs and future predictions. The second proposed architecture, presented in Figure \ref{fig:TSKAN_02}, extends TSMixer by adding a new KAN-based time mixing layer between mixer layers and temporal projection to intensify the capability to uncover the temporal patterns in time series. All architectures use a two-depth KAN layer.

\section{Experimental Results}
In this section, we evaluate the forecasting performance of the two proposed TSKANMixer architectures, presented in Figures \ref{fig:TSKAN_01} and \ref{fig:TSKAN_02}. We evaluate the performance of our proposed TSKANMixer on commonly used benchmark datasets of multivariate time series that have no missing values and equal lengths across all series: Electricity Transformer Temperature (ETT) long-term forecasting dataset, introduced by Zhou et al. \cite{zhou2021informer}, NN5 forecasting competition dataset \cite{taieb2012review}, Computational Intelligence in Forecasting (CIF) 2016 forecasting competition dataset \cite{vstvepnivcka2017results}, FRED-MD dataset \cite{mccracken2016fred}, Exchange dataset\cite{lai2018modeling} and Hospital dataset \cite{hyndman2008forecasting}.

\paragraph{Datasets}
The Electricity Transformer Temperature is a crucial indicator in the electric power long-term deployment. This dataset consists of two years of data from two separate counties in China. Each dataset includes the target variable ``oil temperature'' (OT) and six power load features \cite{zhou2021informer}. We use publicly available data that have been pre-processed by Wu et al. \cite{wu2021autoformer}. The NN5 dataset contains 111 time series of daily cash withdrawals from Automated Teller Machines (ATM) in the UK \cite{godahewa2021monash}. The Computational Intelligence in Forecasting (CIF) 2016 contains 72 monthly time series. Out of these series, 24 series originate from the banking sector, and the remaining 48 series are artificially generated \cite{godahewa2021monash}. In this paper, we use only the 48 series of equal length. The Hospital dataset collects 767 monthly time series showing patient counts related to medical products from January 2000 to December 2006 \cite{godahewa2021monash}. The Exchange dataset is the collection of the daily exchange rates of eight foreign countries, including Australia, Britain, Canada, Switzerland, China, Japan, New Zealand and Singapore, ranging from 1990 to 2016 \cite{lai2018modeling}. The FRED-MD dataset contains 107 monthly time series showing a set of macro-economic indicators from the Federal Reserve Bank \cite{mccracken2016fred}. Each dataset is standardized to achieve zero-mean normalization to ensure a fair comparison with TSMixer \cite{chen2023tsmixer}. We split the data to ensure that the test set's size closely matches the prediction length, maximizing the amount of data available for training. The statistics of the benchmark datasets and data splits are presented in Table \ref{table:ts_dataset}.

\begin{table}[htbp]
  \caption{Time Series Forecasting Datasets}
  \label{table:ts_dataset}
  \scriptsize
  \centering
  \begin{adjustbox}{width=1.0\columnwidth}
  \begin{tabularx}{1.0\columnwidth}{@{} >{\hsize=0.35\hsize\centering\arraybackslash}X | >{\hsize=0.15\hsize\centering\arraybackslash}X >{\hsize=0.2\hsize\centering\arraybackslash}X  >{\hsize=0.3\hsize\centering\arraybackslash}X  >{\hsize=0.45\hsize\centering\arraybackslash}X @{}}
    \toprule
    \textbf{dataset} & features & time steps & time granularity & data split (train/valid/test) \\
    \midrule
    ETTh1/h2    & 7 & 17,420 & 1 hour & 12/4/4 month \\
    ETTm1/m2    & 7 & 699,680 & 15 min  & 12/4/4 month \\
    NN5\_daily  & 111 & 791 &  1 day  & 672/59/59 \\
    NN5\_weekly & 111 & 113 & 1 month  & 96/8/8 \\
    CIF\_2016   & 48 & 120 & 1 month  & 96/12/12 \\
    Hospital    & 767 & 84 & 1 month  & 58/12/12 \\
    Exchange    & 8 & 7,588 & 1 day  & 6829/379/379 \\
    FRED\_MD    & 107 & 728 & 1 month  & 698/14/14 \\
    \bottomrule
  \end{tabularx}
  \end{adjustbox}
\end{table}

\paragraph{Experimental Setup}
We focus on evaluating the impact of the KAN layer on TSMixer by comparing it to the original architecture. Thus, we follow the experimental settings in the TSMixer research \cite{chen2023tsmixer} for ETT datasets about data split and hyperparameters. We set the input length $L = 512$ as suggested in Chen et al. \cite{chen2023tsmixer} and evaluate the results for a forecast horizon of $H = 96$. For TSKANMixer's hyperparameters on ETT, we employ a shallower architecture with fewer mixer blocks and a larger batch size compared to TSMixer. Specifically, while TSMixer uses 4 or 6 mixer blocks, TSKANMixer employs only 2 blocks. Similarly, the batch size differs significantly: 32 for TSMixer and 320 for TSKANMixer. To utilize PyKAN \cite{liu2024kan}, which is implemented in PyTorch, we converted TSMixer's TensorFlow code to PyTorch to implement TSKANMixer. We verified the code conversion by comparing the results with those reported in the original TSMixer paper \cite{chen2023tsmixer} using the ETT dataset, as shown in Table \ref{table:tsmixer} in the Appendix.

In addition, we extensively perform experiments on various publicly available datasets that were not included in the original TSMixer paper \cite{chen2023tsmixer}. We conduct a grid search for TSMixer on the hyperparameter spaces: batch size = $\{8, 16, 32\}$, mixer blocks = $\{2, 4, 6\}$, dropout = $\{0.3, 0.5, 0.7, 0.9\}$, feature hidden size = $\{8, 16, 32, 64\}$, and learning rate = $\{0.0001, 0.001\}$. The models are trained for 1000 epochs with proper early stopping. We select the best configuration of TSMixer for the results shown in Table \ref{table:result_01}. For TSKANMixer's hyperparameters, we conducted manual exploration with limited parameter combinations, as an exhaustive grid search was computationally prohibitive due to the larger parameter space introduced by KAN parameters (e.g., B-spline grids, order of B-spline, and KAN hidden size). Training is also limited to 200 epochs with strict early stopping for the extended datasets. Further details on hyperparameters are summarized in Table \ref{table:hyerparamter_tsmixer} in the Appendix.

As benchmark comparisons, we select various state-of-the-art time series models including MLP-based Series-core Fused Time Series (SOFTS) \cite{han2024softs}, MLP-based TimeMixer \cite{wang2024timemixer}, GNN-based Spectral Temporal Graph Neural Network (StemGNN) \cite{cao2020spectral}, Transformer-based Informer \cite{zhou2021informer}, and Simple MLP for multivariate forecasting. All of these models use the same prediction length ($H$) and input length ($L$) for each dataset as we do for TSMixer \cite{chen2023tsmixer}. We calculate mean squared error (MSE) and mean absolute error (MAE) as the evaluation metrics. We minimize the mean square error (MSE) or the mean absolute error (MAE) as a loss function and evaluate it over a forecast horizon. All models were trained and tested on an ml.g4dn.xlarge GPU instance, powered by a single NVIDIA T4 GPU with 16GB memory.

\paragraph{Experiments}
We evaluate two versions of TSKANMixer proposed in Figure \ref{fig:TSKAN} on popular multivariate forecasting benchmark datasets, comparing them against TSMixer and other state-of-the-art time series models. Table \ref{table:result_01} summarizes the comprehensive comparison of 8 time series forecasting models across 10 datasets using MSE and MAE metrics. The top three results for each dataset are highlighted in bold, with the best performance underlined.

Overall, the evaluation results in Table \ref{table:result_01} show that no time series forecasting model dominantly outperforms others across all datasets. Among benchmark models, TSMixer and SOFTS demonstrate relatively better performance than other models, followed closely by Informer, while TimeMixer shows moderate performance. MLP and StemGNN exhibit lower accuracy. Notably, StemGNN encounters an out-of-memory issue on the ETT dataset. As a result, StemGNN's performance on the ETT dataset is not reported in Table \ref{table:result_01}.

The performance improvements of TSKANMixer models compared to TSMixer are indicated by percentage changes ($\Delta\%$) under TSKANMixer in Table \ref{table:result_01}. For instance, on the ETTh2 dataset, TSKANMixer (v02) shows a substantial 18.97\% improvement in MSE and 9.41\% in MAE over the TSMixer. The performance improvements from TSMixer show that the predictions obtained by one of TSKANMixer are better than the baseline TSMixer by Chen et al. \cite{chen2023tsmixer} in MSE or MAE across eight datasets, except for CIF 2016 and FRED-MD. In particular, TSKANMixer demonstrates the best or second-best performance on ETTh1, ETTh2, ETTm1, ETTm2, NN5 daily, NN5 weekly, Hospital, and FRED-MD. Both versions of TSKANMixer achieved a top-three ranking 7 times each out of 10 datasets. The result implies that the KAN layer improves prediction performance over the original TSMixer architecture. As an exception, in the CFI 2016 case, all models show poor performance on multivariate predictions, showing significantly high MSE on the normalized dataset. Only StemGNN shows the best performance, and it is the only dataset where StemGNN ranks in the top three. This could imply that the dataset has different time series characteristics that are not captured by current variants of TSKANMixer and TSMixer.

\begin{table*}[htbp]
  \caption{Evaluation results on the public time-series datasets}
  \label{table:result_01}
  \scriptsize
  \centering
  \makebox[\textwidth]{
  \begin{adjustbox}{width=1.0\textwidth}
  \begin{tabularx}{1\textwidth}{@{} >{\hsize=0.65\hsize\centering\arraybackslash}X >{\hsize=0.1\hsize\centering\arraybackslash}X >{\hsize=0.1\hsize\centering\arraybackslash}X | >{\hsize=0.71\hsize\centering\arraybackslash}X >{\hsize=0.71\hsize\centering\arraybackslash}X | >{\hsize=0.71\hsize\centering\arraybackslash}X >{\hsize=0.71\hsize\centering\arraybackslash}X | >{\hsize=0.25\hsize\centering\arraybackslash}X >{\hsize=0.3\hsize\centering\arraybackslash}X | >{\hsize=0.25\hsize\centering\arraybackslash}X >{\hsize=0.3\hsize\centering\arraybackslash}X | >{\hsize=0.25\hsize\centering\arraybackslash}X >{\hsize=0.3\hsize\centering\arraybackslash}X | >{\hsize=0.25\hsize\centering\arraybackslash}X >{\hsize=0.3\hsize\centering\arraybackslash}X | >{\hsize=0.25\hsize\centering\arraybackslash}X >{\hsize=0.3\hsize\centering\arraybackslash}X | >{\hsize=0.25\hsize\centering\arraybackslash}X >{\hsize=0.3\hsize\centering\arraybackslash}X @{}}
    \toprule
    & & & \multicolumn{2}{>{\hsize=1.72\hsize\centering\arraybackslash}X}{TSKANMixer (v01)} & \multicolumn{2}{>{\hsize=1.72\hsize\centering\arraybackslash}X}{TSKANMixer (v02)} & \multicolumn{2}{>{\hsize=0.8\hsize\centering\arraybackslash}X}{TSMixer} & \multicolumn{2}{>{\hsize=0.8\hsize\centering\arraybackslash}X}{SOFTS} & \multicolumn{2}{>{\hsize=0.8\hsize\centering\arraybackslash}X}{TimeMixer} & \multicolumn{2}{>{\hsize=0.8\hsize\centering\arraybackslash}X}{MLP} & \multicolumn{2}{>{\hsize=0.8\hsize\centering\arraybackslash}X}{StemGNN} & \multicolumn{2}{>{\hsize=0.8\hsize\centering\arraybackslash}X}{Informer} \\
    \cmidrule(lr){4-5} \cmidrule(lr){6-7} \cmidrule(lr){8-9} \cmidrule(lr){10-11} \cmidrule(lr){12-13} \cmidrule(lr){14-15} \cmidrule(lr){16-17} \cmidrule(lr){18-19}
    \textbf{dataset} & L & H & MSE ($\Delta$\%) & MAE ($\Delta$\%) & MSE ($\Delta$\%) & MAE ($\Delta$\%) & MSE & MAE & MSE & MAE & MSE & MAE & MSE & MAE & MSE & MAE & MSE & MAE \\
    \midrule
    ETTh1 & 512 & 96 & \textbf{\underline{0.285}} (33.57\%) & 0.398 (2.69\%)  & \textbf{0.296} (31.00\%) & 0.405 (0.98\%) & 0.429 & 0.409 & 0.609 & 0.441 & 0.516 & 0.415  & 0.517 & 0.595 & - & - & \textbf{0.337} & 0.403 \\
    ETTh2 & 512 & 96 & 0.199 (-2.05\%) & 0.334 (1.76\%) & \textbf{0.158} (18.97\%) & 0.308 (9.41\%) & \textbf{0.195} & 0.340 & \textbf{\underline{0.135}} & 0.277 & 0.593 & 0.552  & 0.418 & 0.529 & - & - & 0.208 & 0.373 \\
    ETTm1 & 512 & 96 & \textbf{\underline{0.190}} (34.26\%) & 0.296 (13.20\%) & 0.281 (2.77\%) & 0.348 (-2.05\%) & 0.289 & 0.341 & \textbf{0.211} & 0.307 & 0.271 & 0.355  & 0.406 & 0.378 & - & - & \textbf{0.259} & 0.377 \\
    ETTm2 & 512 & 96 & \textbf{0.131} (9.66\%)  & 0.268 (3.60\%)  & \textbf{0.109} (24.83\%) & 0.251 (9.71\%) & 0.145 & 0.278 & 0.148 & 0.279 & \textbf{\underline{0.108}} & 0.245  & 0.240 & 0.398 & - & - & 0.215 & 0.362 \\
    NN5\_daily & 56 & 56 & \textbf{0.521} (-1.36\%) & 0.498 (-1.01\%) & \textbf{\underline{0.506}} (1.56\%) & 0.485 (1.62\%) & \textbf{0.514} & 0.493 & 0.545 & 0.506 & 0.627 & 0.582  & 0.641 & 0.582 & 0.561 & 0.515 & 0.544 & 0.521 \\
    NN5\_weekly & 16 & 8 & \textbf{\underline{0.878}} (2.34\%) & 0.731 (1.08\%) & \textbf{0.897} (0.22\%) & 0.736 (0.41\%) & 0.899 & 0.739 & 0.938 & 0.771 & \textbf{0.901} & 0.736  & 1.195 & 0.883 & 1.758 & 1.014 & 1.177 & 0.859 \\
    CIF\_2016 & 24 & 12 & 3.631 (-34.58\%) & 1.026 (-31.37\%) & 2.936 (-8.82\%) & 0.895 (-14.59\%) & \textbf{2.698} & 0.781 & \textbf{2.585} & 0.687 & 3.736 & 0.934  & 4.963 & 1.241 & \textbf{\underline{2.475}} & 0.760 & 5.275 & 1.385 \\
    Hospital & 24 & 12 & \textbf{1.429} (11.08\%) & 0.928 (6.64\%) & 1.556 (3.17\%) & 0.979 (1.51\%) & 1.607 & 0.994 & \textbf{\underline{1.338}} & 0.875 & 1.525 & 0.939  & \textbf{1.454} & 0.939 & 1.475 & 0.929 & 1.784 & 1.031 \\
    Exchange & 60 & 30 & 0.017 (5.56\%) & 0.099 (7.47\%) & \textbf{0.016} (11.11\%) & 0.094 (12.15\%) & 0.018 & 0.107 & \textbf{\underline{0.011}} & 0.084 & 0.025 & 0.115  & 0.139 & 0.295 & 1.802 & 1.060 & \textbf{0.015} & 0.088 \\
    FRED\_MD & 48 & 12 & \textbf{0.037} (-5.71\%) & 0.133 (-6.4\%) & \textbf{0.036} (-2.86\%) & 0.125 (0\%) & \textbf{\underline{0.035}} & 0.125 & 0.052 & 0.122 & 0.046 & 0.126  & 0.126 & 0.255 & 0.101 & 0.202 & 0.049 & 0.145 \\
    \bottomrule
  \end{tabularx}
  \end{adjustbox}
  }
\end{table*}

On the other hand, TSKANMixer exhibits significantly slower training times compared to TSMixer due to the incorporation of the KAN layer. According to PyKAN \cite{liu2024kan}, the primary bottleneck of KAN is its slow training process, as KANs introduce additional complexity and computations. The study reports that KANs are typically 10 times slower than MLPs, given the same number of parameters. The training limitation constrains the testing of TSKANMixer on larger datasets and hinders extensive hyperparameter tuning in this paper. In addition, TSKANMixer sometimes requires more epochs to complete training than TSMixer on ETT datasets. As a result, there is a case that TSKANMixer's training time is approximately up to 50 times slower than that of the original TSMixer as shown in Table \ref{table:result_02}.

\begin{table*}[htbp]
  \caption{Computational Time}
  \label{table:result_02}
  \small
  \centering
  \makebox[\textwidth]{
  \begin{adjustbox}{width=0.9\textwidth}
  \begin{tabularx}{1\textwidth}{@{} >{\hsize=0.4\hsize\centering\arraybackslash}X >{\hsize=0.1\hsize\centering\arraybackslash}X >{\hsize=0.1\hsize\centering\arraybackslash}X | >{\hsize=0.4\hsize\centering\arraybackslash}X >{\hsize=0.45\hsize\centering\arraybackslash}X | >{\hsize=0.4\hsize\centering\arraybackslash}X >{\hsize=0.45\hsize\centering\arraybackslash}X | >{\hsize=0.4\hsize\centering\arraybackslash}X >{\hsize=0.45\hsize\centering\arraybackslash}X @{}}
    \toprule
    & & & \multicolumn{2}{>{\hsize=0.95\hsize\centering\arraybackslash}X}{TSKANMixer (v01)} & \multicolumn{2}{>{\hsize=0.95\hsize\centering\arraybackslash}X}{TSKANMixer (v02)} & \multicolumn{2}{>{\hsize=0.95\hsize\centering\arraybackslash}X}{TSMixer}\\
    \cmidrule(lr){4-5} \cmidrule(lr){6-7} \cmidrule(lr){8-9}
    \textbf{dataset} & L & H & time/epoch (sec) & training time (sec) & time/epoch (sec) & training time (sec) & timee/poch (sec) & training time (sec) \\
    \midrule
    ETTh1 & 512 & 96 & 21.29 & 4885.08  & 63.70 & 11869.71 & 4.10 & 263.01 \\
    ETTh2 & 512 & 96 & 40.69 & 12276.67 & 63.59 & 19302.69 & 4.74 & 312.52 \\
    ETTm1 & 512 & 96 & 88.81 & 11662.25 & 264.63 & 39819.87 & 26.83 & 1282.46 \\
    ETTm2 & 512 & 96 & 171.43 & 19242.63 & 263.53 & 63953.10 & 27.26 & 2032.07 \\
    NN5\_daily & 56 & 56 & 25.82 & 1884.76 & 22.80 & 1869.92 & 0.62 & 135.86 \\
    NN5\_weekly & 16 & 8 & 6.19 & 1354.61 & 3.21 & 125.01 & 0.08 & 28.89 \\
    CIF\_2016 & 24 & 12 & 2.97 & 204.98 & 4.03 & 454.99 & 0.07 & 59.96 \\
    Hospital & 24 & 12 & 20.32 & 589.23  & 15.09 & 2489.44 & 0.04 & 11.9 \\
    Exchange & 60 & 30 & 5.04 & 181.33 & 9.70 & 339.55 & 10.18 & 2421.82 \\
    FRED\_MD & 48 & 12 & 30.89 & 1730.17 & 19.98 & 3036.34 & 0.33 & 181.19 \\
    \bottomrule
  \end{tabularx}
  \end{adjustbox}
  }
\end{table*}

\begin{figure}[htbp]
\centering
\begin{subfigure}[b]{0.8\columnwidth}
   \centering
   \includegraphics[width=\textwidth]{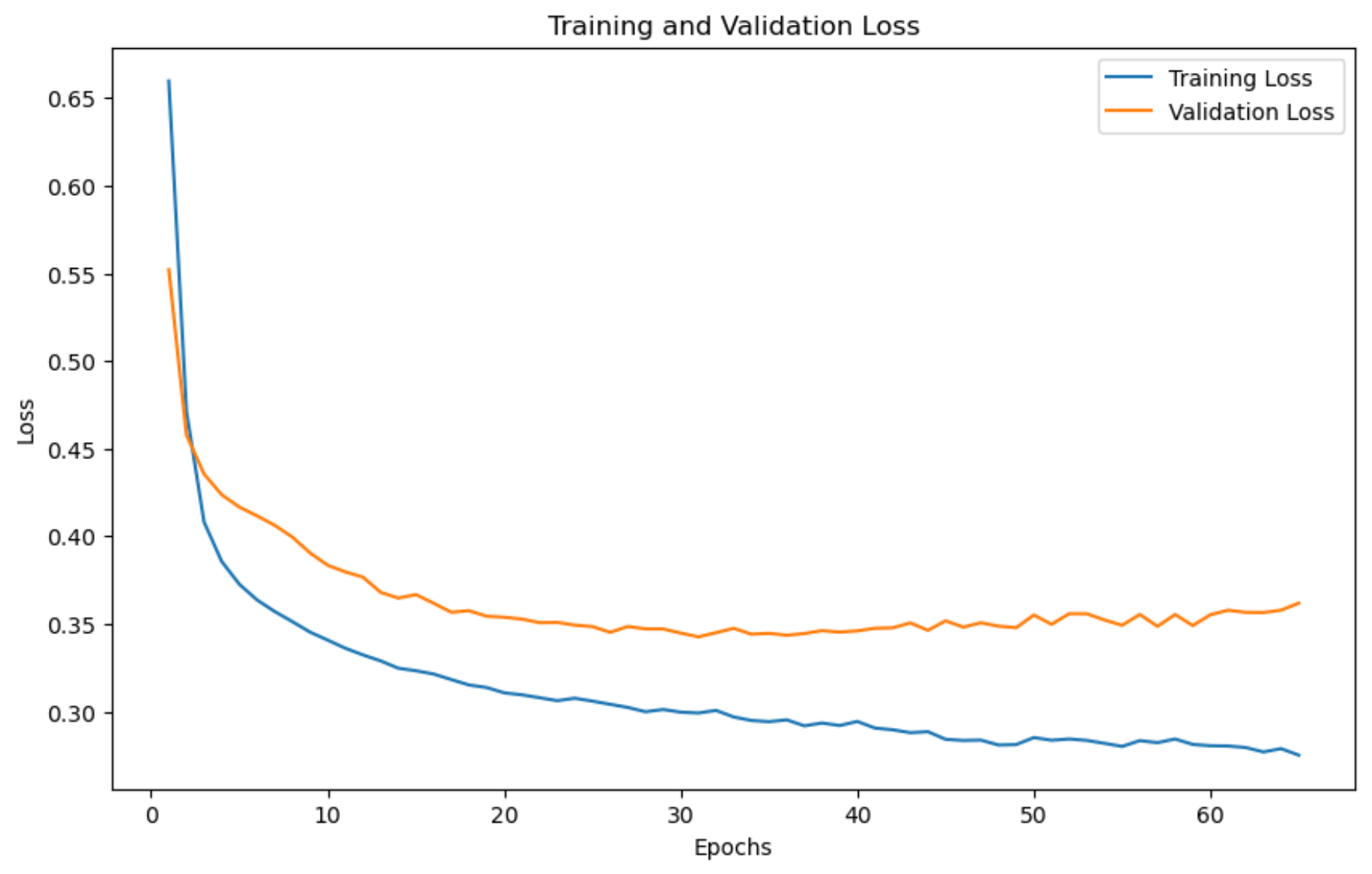}
   \caption{TSMixer}
   \label{fig:plot_01}
\end{subfigure}
\hfill %
\begin{subfigure}[b]{0.8\columnwidth}
   \centering
   \includegraphics[width=\textwidth]{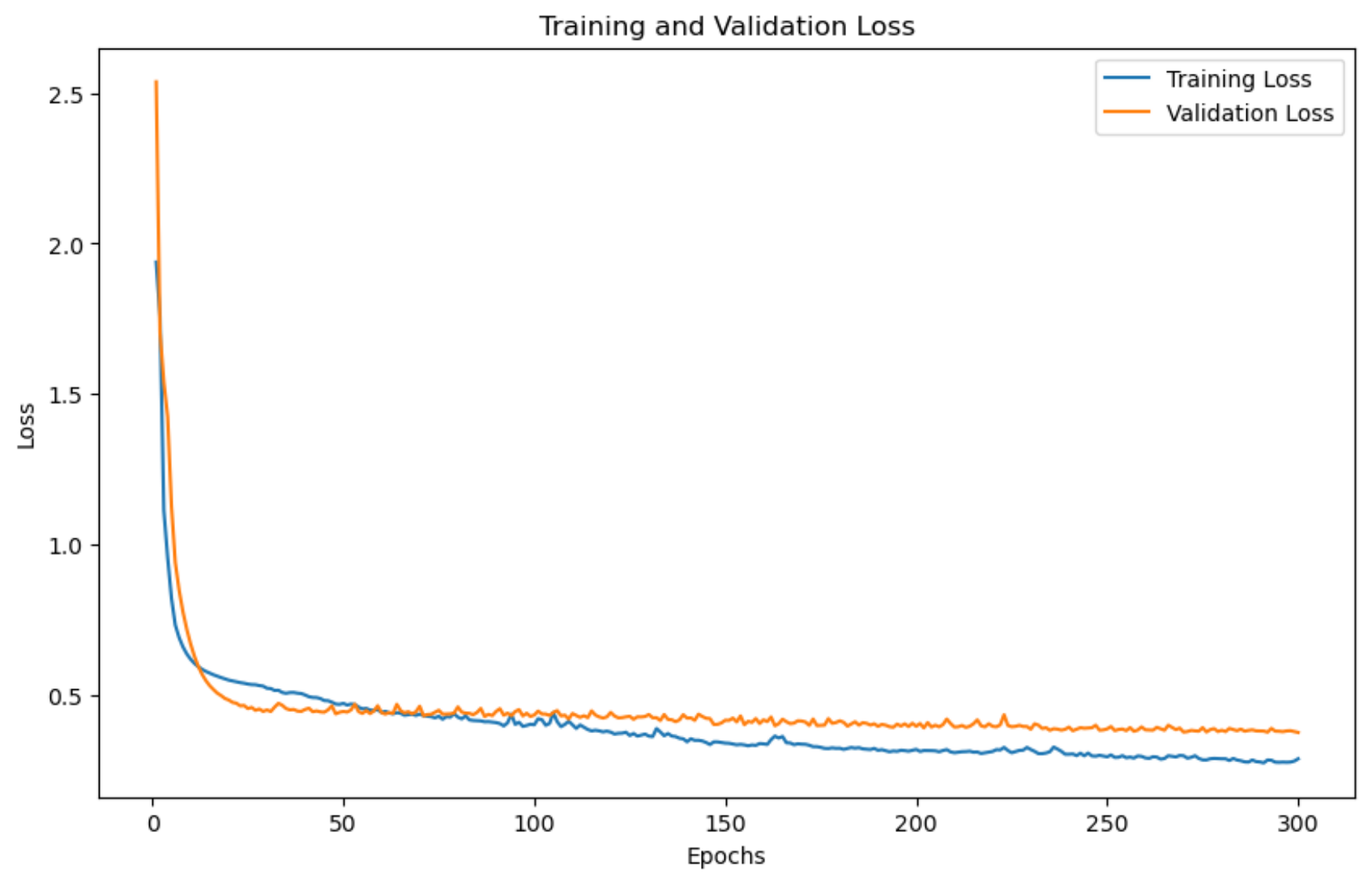}
   \caption{TSKANMixer}
   \label{fig:plot_02}
\end{subfigure}
\caption{Training and validation over epochs on ETTh2}
\label{fig:plot}
\end{figure}

To illustrate the slow training process, we visualize the training and validation losses over the training epochs for TSMixer and TSKANMixer, as shown in Figure \ref{fig:plot}. On the ETT datasets, TSMixer starts with a relatively low initial loss value compared to TSKANMixer. It reaches the best epoch at an earlier stage (e.g., less than 50 epochs) and starts overfitting afterwards. This is shown by the increasing validation loss and the divergence between its training loss and validation loss as the number of epochs increases in Figure \ref{fig:plot_01}. On the other hand, TSKANMixer shows a poor initial loss value, but it steadily decreases the validation loss as the number of epochs increases without overfitting quickly (e.g., the best epoch happens after 50 epochs), as shown in Figure \ref{fig:plot_02}. TSKANMixer effectively captures the underlying generalized patterns present in the data, rather than falling into local optima. 

\section{Conclusion and Future Work}
In this paper, we explored the application of KAN to the TSMixer model for time series forecasting and introduced two variants of TSKANMixer. We demonstrate that the TSKANMixer models generally improve prediction performance over the original TSMixer models. This improvement is achieved by either replacing the fully-connected layer with KAN in temporal projection or by adding a time-mixing layer with KAN. However, we also note that the KAN layer slows down the training process. This work highlights the promising application of KANs in time series analysis. We hope these results provide insights for future research on KAN for time-series forecasting models to improve the capability to capture complex patterns in time series data.

Future work could explore improving the training time for a generalized architecture having wider and deeper KANs beyond the current two-layer model. Additionally, developing a more efficient KAN implementation would facilitate comprehensive hyperparameter tuning on TSKANMixer, potentially unlocking the full potential of KAN-based models. Finally, further exploiting the interpretability and robustness of KAN-based models through symbolic regression could open opportunities to develop more effective and efficient time series models.

\bibliography{main}

\appendix
\section{Appendix A. TSMixer Implemenation}
\label{appendix:tsmixer}

\begin{table}[htbp]
  \caption{TSMixer Comparison on ETT}
  \label{table:tsmixer}
  \centering
  \begin{adjustbox}{width=0.9\columnwidth}
  \begin{tabularx}{1.0\columnwidth}{@{} >{\hsize=0.3\hsize\centering\arraybackslash}X >{\hsize=0.1\hsize\centering\arraybackslash}X >{\hsize=0.1\hsize\centering\arraybackslash}X | >{\hsize=0.35\hsize\centering\arraybackslash}X >{\hsize=0.35\hsize\centering\arraybackslash}X | >{\hsize=0.35\hsize\centering\arraybackslash}X  >{\hsize=0.35\hsize\centering\arraybackslash}X @{}}
    \toprule
    & & & \multicolumn{2}{>{\hsize=0.8\hsize\centering\arraybackslash}X}{TSMixer (TensorFlow) } & \multicolumn{2}{>{\hsize=0.8\hsize\centering\arraybackslash}X}{TSMixer (PyTorch)} \\
    \cmidrule(lr){4-5} \cmidrule(lr){6-7}
    \textbf{dataset} & L & H & MSE & MAE & MSE & MAE \\
    \midrule
    ETTh1 & 512 & 96 & 0.361 & 0.392 & 0.429 & 0.409  \\
    ETTh2 & 512 & 96 & 0.274 & 0.341 & 0.195 & 0.340  \\
    ETTm1 & 512 & 96 & 0.285 & 0.339 & 0.289 & 0.341  \\
    ETTm2 & 512 & 96 & 0.163 & 0.252 & 0.145 & 0.278  \\
    \bottomrule
  \end{tabularx}
  \end{adjustbox}
\end{table}

\section{Appendix B. Forecasted Values Visualization}
\label{appendix:visualization}

\begin{figure}[htbp]
\centering
\begin{subfigure}[b]{1.0\columnwidth}
   \centering
   \includegraphics[width=\textwidth]{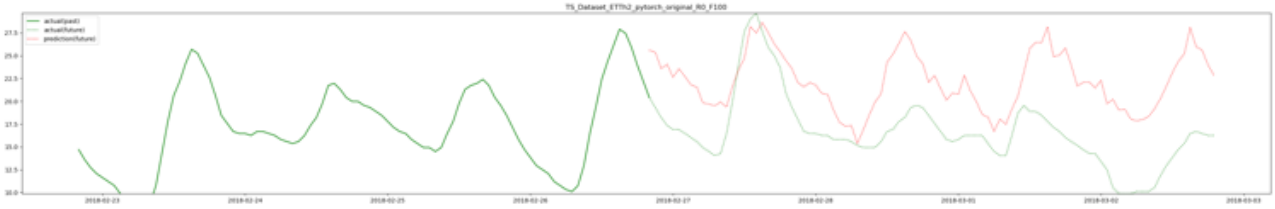}
   \caption{Forecasted values (H=96) using TSMixer (MAE = 0.340) on ETTh2}
   \label{fig:ts_plot_01}
\end{subfigure}
\vskip\baselineskip %
\begin{subfigure}[b]{1.0\columnwidth}
   \centering
   \includegraphics[width=\textwidth]{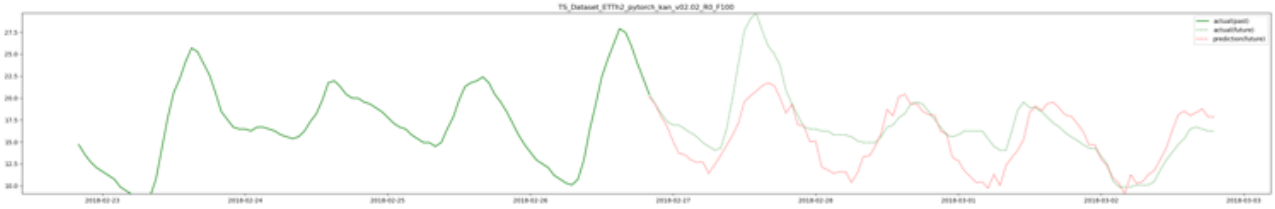}
   \caption{Forecasted values (H=96) using TSKANMixer v02 (MAE = 0.327) on ETTh2}
   \label{fig:ts_plot_02}
\end{subfigure}
\caption{Visualization of predictions: true (green) and forecasted (red) values for the target (OT) feature}
\end{figure}

\section{Appendix C. Hyperparameters}
\label{appendix:hyperparameters}

\begin{table*}[htbp]
  \caption{Hyerparamter configurations for TSMixer}
  \label{table:hyerparamter_tsmixer}
  \small
  \centering
  \makebox[\textwidth]{
  \begin{adjustbox}{width=0.9\textwidth}
  \begin{tabularx}{1\textwidth}{@{} >{\hsize=0.3\hsize\centering\arraybackslash}X >{\hsize=0.1\hsize\centering\arraybackslash}X >{\hsize=0.1\hsize\centering\arraybackslash}X | >{\hsize=0.5\hsize\centering\arraybackslash}X >{\hsize=0.5\hsize\centering\arraybackslash}X >{\hsize=0.55\hsize\centering\arraybackslash}X >{\hsize=0.5\hsize\centering\arraybackslash}X >{\hsize=0.55\hsize\centering\arraybackslash}X @{}}
    \toprule
    & & & \multicolumn{5}{>{\hsize=3.0\hsize\centering\arraybackslash}X}{TSMixer}\\
    \cmidrule(lr){4-8}
    \textbf{dataset} & L & H & Batch & Blocks & Dropout & Hidden size & Learing rate \\
    \midrule
    ETTh1 & 512 & 96 & 32 & 2  & 0.3 & 64 & 0.0001 \\
    ETTh2 & 512 & 96 & 32 & 4  & 0.3 & 64 & 0.0001 \\
    ETTm1 & 512 & 96 & 32 & 6  & 0.9 & 16 & 0.0001 \\
    ETTm2 & 512 & 96 & 32 & 6  & 0.3 & 16 & 0.0001 \\
    NN5\_daily & 56 & 56 & 16 & 6 & 0.3 & 64 & 0.001 \\
    NN5\_weekly & 16 & 8 & 16 & 6 & 0.9 & 64 & 0.001  \\
    CIF\_2016 & 24 & 12 & 8 & 4 & 0.9 & 8 & 0.001 \\
    Hospital & 24 & 12 & 8 & 6 & 0.5 & 16 & 0.001 \\
    Exchange & 60 & 30 & 8 & 6 & 0.5 & 64 & 0.001 \\
    FRED\_MD & 48 & 12 & 32 & 6 & 0.3 & 16 & 0.001 \\
    \bottomrule
  \end{tabularx}
  \end{adjustbox}
  }
\end{table*}

\begin{table*}[htbp]
  \caption{Hyerparamter configurations for TSKANMixer (v01)}
  \label{table:hyerparamter_tskanmixer_01}
  \small
  \centering
  \makebox[\textwidth]{
  \begin{adjustbox}{width=0.9\textwidth}
  \begin{tabularx}{1\textwidth}{@{} >{\hsize=0.4\hsize\centering\arraybackslash}X >{\hsize=0.1\hsize\centering\arraybackslash}X >{\hsize=0.1\hsize\centering\arraybackslash}X | >{\hsize=0.4\hsize\centering\arraybackslash}X >{\hsize=0.4\hsize\centering\arraybackslash}X >{\hsize=0.4\hsize\centering\arraybackslash}X >{\hsize=0.5\hsize\centering\arraybackslash}X >{\hsize=0.5\hsize\centering\arraybackslash}X >{\hsize=0.4\hsize\centering\arraybackslash}X >{\hsize=0.45\hsize\centering\arraybackslash}X >{\hsize=0.4\hsize\centering\arraybackslash}X @{}}
    \toprule
    & & & \multicolumn{8}{>{\hsize=4.0\hsize\centering\arraybackslash}X}{TSKANMixer (v01)}\\
    \cmidrule(lr){4-11}
    \textbf{dataset} & L & H & Batch & Blocks & Dropout & Hidden size & Learing rate & KAN\_dim & KAN\_grid & KAN\_k \\
    \midrule
    ETTh1 & 512 & 96 & 320 & 2  & 0.3 & 64 & 0.0001 & 512 & 5 & 3 \\
    ETTh2 & 512 & 96 & 320 & 2  & 0.3 & 64 & 0.0001  & 1025 & 5 & 3 \\
    ETTm1 & 512 & 96 & 320 & 2  & 0.3 & 64 & 0.0001  & 512 & 5 & 3 \\
    ETTm2 & 512 & 96 & 320 & 4  & 0.3 & 64 & 0.0001  & 1025 & 5 & 3 \\
    NN5\_daily & 56 & 56 & 16 & 4 & 0.3 & 32 & 0.001 & 56 & 10 & 2 \\
    NN5\_weekly & 16 & 8 & 8 & 6 & 0.7 & 111 & 0.001 & 33 & 3 & 3 \\
    CIF\_2016 & 24 & 12 & 16 & 2 & 0.9 & 64 & 0.001 & 12 & 1 & 10 \\
    Hospital & 24 & 12 & 8 & 2 & 0.5 & 767 & 0.001 & 24 & 10 & 2 \\
    Exchange & 60 & 30 & 128 & 4 & 0.3 & 4 & 0.001 & 15 & 10 & 3 \\
    FRED\_MD & 48 & 12 & 32 & 4 & 0.3 & 16 & 0.001 & 12 & 10 & 7\\
    \bottomrule
  \end{tabularx}
  \end{adjustbox}
  }
\end{table*}

\begin{table*}[htbp]
  \caption{Hyerparamter configurations for TSKANMixer (v02)}
  \label{table:hyerparamter_tskanmixer_02}
  \small
  \centering
  \makebox[\textwidth]{
  \begin{adjustbox}{width=0.9\textwidth}
  \begin{tabularx}{1\textwidth}{@{} >{\hsize=0.4\hsize\centering\arraybackslash}X >{\hsize=0.1\hsize\centering\arraybackslash}X >{\hsize=0.1\hsize\centering\arraybackslash}X | >{\hsize=0.4\hsize\centering\arraybackslash}X >{\hsize=0.4\hsize\centering\arraybackslash}X >{\hsize=0.4\hsize\centering\arraybackslash}X >{\hsize=0.5\hsize\centering\arraybackslash}X >{\hsize=0.5\hsize\centering\arraybackslash}X >{\hsize=0.4\hsize\centering\arraybackslash}X >{\hsize=0.45\hsize\centering\arraybackslash}X >{\hsize=0.4\hsize\centering\arraybackslash}X @{}}
    \toprule
    & & & \multicolumn{8}{>{\hsize=4.0\hsize\centering\arraybackslash}X}{TSKANMixer (v02)}\\
    \cmidrule(lr){4-11}
    \textbf{dataset} & L & H & Batch & Blocks & Dropout & Hidden size & Learing rate & KAN\_dim & KAN\_grid & KAN\_k \\
    \midrule
    ETTh1 & 512 & 96 & 320 & 2 & 0.3 & 64 & 0.0001 & 1025 & 5 & 3 \\
    ETTh2 & 512 & 96 & 320 & 2 & 0.3 & 64 & 0.0001 & 1025 & 5 & 3 \\
    ETTm1 & 512 & 96 & 320 & 2 & 0.3 & 64 & 0.0001 & 1025 & 5 & 3 \\
    ETTm2 & 512 & 96 & 320 & 2 & 0.3 & 64 & 0.0001 & 1025 & 5 & 3 \\
    NN5\_daily & 56 & 56 & 16 & 4 & 0.9 & 32 & 0.001 & 14 & 2 & 3 \\
    NN5\_weekly & 16 & 8 & 8 & 6 & 0.7 & 32 & 0.001 & 8 & 7 & 3 \\
    CIF\_2016 & 24 & 12 & 16 & 4 & 0.4 & 24 & 0.001 & 12 & 7 & 3 \\
    Hospital & 24 & 12 & 8 & 6 & 0.3 & 16 & 0.001 & 3 & 10 & 2 \\
    Exchange & 60 & 30 & 32 & 2 & 0.3 & 16 & 0.001 & 15 & 10 & 2 \\
    FRED\_MD & 48 & 12 & 16 & 2 & 0.5 & 16 & 0.001 & 5 & 5 & 2 \\
    \bottomrule
  \end{tabularx}
  \end{adjustbox}
  }
\end{table*}

\end{document}